\documentclass{article}
\usepackage{ijcai18}

\usepackage{times}
\usepackage{xcolor}
\usepackage{soul}
\usepackage[utf8]{inputenc}
\usepackage[small]{caption}
\usepackage{amssymb}
\usepackage{amsmath}
\usepackage{kbordermatrix}
\usepackage{fourier} 
\usepackage{array}
\usepackage{makecell}
\usepackage{algorithm}
\usepackage[noend]{algpseudocode}
\usepackage{graphicx}
\usepackage{pgfplots}
\usepackage{subcaption}
\usepackage{multirow}
\usepackage{booktabs}
\usepackage[export]{adjustbox}
\usepackage[fleqn]{mathtools}
\usepackage{cases}
\usepackage{url}
\usepackage{CJK}

\pgfplotsset{compat=1.14}

\algdef{SE}[VARIABLES]{Variables}{EndVariables}
   {\algorithmicvariables}
   {\algorithmicend\ \algorithmicvariables}
\algnewcommand{\algorithmicvariables}{\textbf{global variables}}

\DeclareMathOperator*{\argmax}{arg\,max}

\title{KNPTC: Knowledge and Neural Machine Translation Powered Chinese Pinyin Typo Correction}

\author{
Hengyi Cai$^1$, 
Xingguang Ji$^2$, 
Yonghao Song$^1$,
Yan Jin$^1$,
Yang Zhang$^3$,
Mairgup Mansur$^3$,
Xiaofang Zhao$^1$
\\ 
$^1$ Institute of Computing Technology, Chinese Academy of Sciences\\
$^2$ Beijing University of Posts and Telecommunications\\
$^3$ Sogou, Inc.  \\
caihengyi@ict.ac.cn,
jixg1994@bupt.edu.cn,
\{songyonghao, jinyan\}@ict.ac.cn,\\
\{zhangyang, maerhufu\}@sogou-inc.com,
zhaoxf@ict.ac.cn
}
\begin{document}

\begin{CJK}{UTF8}{min}

\maketitle

\begin{abstract}
	Chinese pinyin input methods are very important for Chinese language processing.
	Actually, users may make typos inevitably when they input pinyin.
	Moreover, pinyin typo correction has become an increasingly important task with the popularity of 
	smartphones and the mobile Internet.
	How to exploit the knowledge of users typing behaviors and support the typo correction for acronym pinyin remains a challenging problem.
	To tackle these challenges,
	we propose KNPTC, 
	a novel approach based on neural machine translation~(NMT).
	In contrast to previous work, 
	KNPTC is able to integrate explicit knowledge into NMT for pinyin typo correction,
	and is able to learn to correct a variety of typos without guidance of manually selected constraints or language-specific features.
	In this approach, we first obtain the transition probabilities between adjacent letters based on large-scale real-life datasets.
	Then, we construct the ``ground-truth'' alignments of training sentence pairs by utilizing these probabilities.
	Furthermore, these alignments are integrated into NMT to capture sensible pinyin typo correction patterns.
	KNPTC is applied to correct typos in real-life datasets, 
	which achieves 32.77\% increment on average in accuracy rate of typo correction compared against the state-of-the-art system.
\end{abstract}

\section{Introduction}

Chinese pinyin is the official system to transcribe Chinese characters into the Latin alphabet.
Based on this transcription system, pinyin input method\footnote{
Throughout this paper, pinyin input method means sentence-based pinyin input method,
which generates a sequence of Chinese characters upon a sequence of pinyin input.
} has become the dominant method for entering Chinese text into computers in China.

\begin{figure}
  \begin{subfigure}[b]{0.24\textwidth}
    \includegraphics[width=\textwidth]{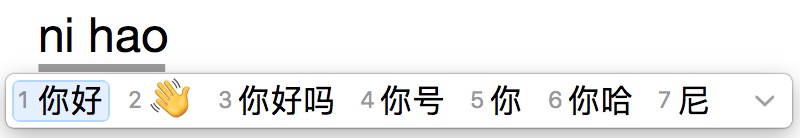}
    \caption{}
    \label{fig:cuoshu1}
  \end{subfigure}
  \begin{subfigure}[b]{0.24\textwidth}
    \includegraphics[width=\textwidth]{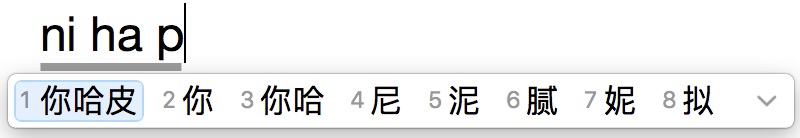}
    \caption{}
    \label{fig:cuoshu2}
  \end{subfigure}
  \caption{pinyin input method for a correct pinyin (a) and a mistyped pinyin (b).}
  \label{fig:cuoshu}
\end{figure}

The typical way to type in Chinese words is in a sequential manner~\cite{DBLP:conf/chi/WangZS01}.
For example, assuming that users want to type in the Chinese word ``你好(Hello)''. 
First, they type in the corresponding pinyin ``nihao'' without delimiters such as “Space” key to segment pinyin syllables. 
Then， a Chinese pinyin input method displays a list of Chinese candidate words which share that pinyin.
Finally, users search the target word from candidates and get the result.
In this way, typing in Chinese words is more complicated than typing in English words.
In fact, pinyin typos have always been a serious problem for Chinese pinyin input methods.
Users often make various typos in the process of inputting. 
As shown in Figure~\ref{fig:cuoshu}, users may mistype ``nihao'' for ``nihap''(the letter \textbf{p} is very close to the letter \textbf{o} on the QWERTY keyboard).
In addition, the user may also fail to input the completely right pinyin simply because he/she is a dialect speaker and does not know the exact pronunciation of the expected Chinese word~\cite{jia2014joint}. 
Moreover, with the boom of smart-phones, pinyin typos are more prone to be made possibly due to the limited size of soft keyboard, and the lack of physical feedback on the touch screen~\cite{jia2014joint}.
When an input method engine~(IME) fails to correct the typo and produce the desired Chinese sentence, the user have to spend extra effort to move the cursor back to the typo and correct it, which results in a very poor user experience~\cite{W13-4416}.
Hence Chinese pinyin typo correction has a significant impact on IME performance.

However, due to the peculiar characteristics of Chinese pinyin, 
typo correction for Chinese pinyin is quite different from that in other languages.
For example, when users want to input ``早上好(Good morning)'', 
they will type ``zaoshanghao'' instead of segmented pinyin sequence ``zao shang hao''. 
Besides, acronym pinyin input\footnote{
Acronym pinyin input means using the first letters of a syllable,
e.g., ``zsh'' for ``早上好''.
}
is very common in 
Chinese daily input.
How to support acronym pinyin input for typo correction is also a challenging problem~\cite{DBLP:conf/ijcai/ZhengLS11}.
For the above reasons, it is impractical to adopt previous typo correction methods for English to pinyin typo correction directly.

Meanwhile, machine translation is well established in the field of automatic error correction(AEC)~\cite{DBLP:conf/emnlp/Junczys-Dowmunt16}. 
With the recent successful work on neural machine translation, neural encoder-decoder models have been used in the AEC as well~\cite{DBLP:conf/emnlp/Junczys-Dowmunt16,DBLP:journals/corr/XieAAJN16}, and have achieved promising results. 
In this paper, we explore new neural models to tackle the unique challenges of Chinese pinyin typo correction.

The main component of neural machine translation~(NMT) systems is a sequence-to-sequence (S2S) model which encodes a sequence of source words into a vector and then generates a sequence of target words from the vector~\cite{DBLP:conf/acl/JiWTGTG17}.
Among different variants of NMT,
attentional NMT~\cite{DBLP:journals/corr/BahdanauCB14,DBLP:conf/emnlp/LuongPM15} 
becomes prevalent due to its ability to use the most relevant part of the source statement in each translation step.
Unlike the classification-based and statistics-based approaches, NMT is more capable of capturing global contextual information, which is crucial to pinyin typo correction. However, due to the characteristics of Chinese pinyin,
to achieve the best performance on pinyin typo correction, 
we still need to improve the standard NMT model to address several challenges: 

\begin{itemize}
	\item First, due to the limited size of soft keyboard on smart-phones, 
	letters close to each other on a Latin keyboard or with similar pronunciations are more likely to be mistyped,
	which leads to a great proportion of pinyin spelling mistakes~\cite{DBLP:conf/acl/ZhengXLSZR11}.
	How to solve these typos caused by the proximity keys is crucial for improving the performance of IMEs.
	\item Second, since acronym pinyin is widely used in Chinese daily input, 
	typo correction for acronym pinyin input greatly affects the user input experience.
	However, due to its shortness and sparsity, 
	previous pinyin typo correction methods failed to support the acronym pinyin input.
	
	\item Third, in addition to the aforementioned task-specific challenges,
	the attention quality of NMT is still unsatisfactory. 
	Lack of explicit knowledge may lead to attention faults and generate inaccurate translations~\cite{DBLP:conf/acl/ZhangWLZ17}.
\end{itemize}

On the one hand, there is a need to introduce prior knowledge into the attention mechanism to improve the performance of NMT. On the other hand, valuable input patterns can be extracted from the user input logs to guide the pinyin typo correction.
In order to overcome the aforementioned challenges, in this paper, we propose an approach called ``\textbf{KNPTC}'', which stands for 
``\textbf{K}nowledge and \textbf{N}eural machine translation powered Chinese \textbf{P}inyin \textbf{T}ypo \textbf{C}orrection''.
KNPTC is able to integrate discrete, probabilistic keyboard neighborhoods information about users pinyin input as prior knowledge into attentional NMT 
to capture more sensible typo correction patterns.
Specifically, we first use large-scale real-life datasets to obtain the transition probabilities between adjacent letters in advance. 
We then construct the ``ground-truth'' alignments of training sentence pairs by utilizing these probabilities.
Finally, distance between the NMT attentions and the ``ground-truth'' alignments is computed as a part of the loss function and need to be minimized in the training procedure.
Considering that the transition probabilities of adjacent letters provide higher quality knowledge about real-life users' input behaviors,
it is expected that the attentional NMT with prior knowledge will achieve better performance on end-to-end pinyin typo correction task.

We find that KNPTC has the following properties:
1) KNPTC can obtain the optimal segmentation and typo correction jointly on the user's original input pinyin sequence;
2) KNPTC is capable of dealing with typos relating to acronym pinyin input.
In the experimental study, 
we compared KNPTC with the Google Input Tools\footnote{\url{https://www.google.com/inputtools/try/}}~(henceforth referred to as \textbf{GoogleIT}) 
and the joint graph model proposed in~\cite{jia2014joint}(henceforth referred to as \textbf{JGM}).
We conduct experiments on real-life datasets and found that 
KNPTC can achieve significantly better performance than the other two baseline systems.

The contributions of this paper are summarized as follows.

\begin{itemize}
\item We propose a solution of exploiting priori knowledge of users' input behaviors to seamlessly integrate the transition probabilities between adjacent letters with attentional NMT for Chinese pinyin typo correction.
\item Compared with previous work, 
KNPTC is more capable of typo correction for acronym pinyin, which is widely used for Chinese people's daily input.
\item We conduct extensive experiments on real-life datasets.
The results show that KNPTC significantly outperforms previous systems(such as GoogleIT and JGM) and is efficient enough for practical use.
\end{itemize}


\section{Related Work}


Chen and Lee first tried to resolve the problem of Chinese pinyin typo correction~\cite{DBLP:conf/acl/ChenL00}.
They designed a statistical typing model to correct the pinyin typos and then used a language model to convert a pinyin sequence to a Chinese characters sequence. 
However, their method can only model a few rules and thus only a limited number of errors can be corrected.
Zheng et al.~proposed an error-tolerant pinyin input method called CHIME using a noisy channel model and language-specific features~\cite{DBLP:conf/ijcai/ZhengLS11}.
Their method assumed that the user inputted pinyin sequence had already been properly segmented by the user, 
which is inconsistent with the situation in real life.
Our model discards this assumption to make it more practical.
Jia and Zhao proposed a joint graph model to find the global optimum of pinyin to Chinese conversion and 
typo correction for the input method engine, and achieved better results than previous studies~\cite{jia2014joint}.
In this paper, we employ this joint graph model as one of our baseline systems.

In addition, Chen et al.~attempted to integrate neural network language models into pinyin IME~\cite{DBLP:conf/paclic/ChenZW15}.
Their main purpose is to improve the decoding predictive performance of IMEs under the premise of ensuring the response speed.
However, due to lack of the ability of typo correction, the model they proposed is not convenient to use in real-life situations.

As for automatic error correction in English, most earlier work~\cite{gao2010large,rozovskaya2010generating} made use of a lexicon 
that contains well-spelled words or context features of words.
Recently, machine translation is well established in the field of automatic error correction in English~\cite{DBLP:conf/emnlp/Junczys-Dowmunt16,DBLP:journals/corr/XieAAJN16}. 
In this work, we mainly focus on the typo correction of Chinese pinyin, which is quite different from other languages for the reasons mentioned before.

\section{Our Approach: KNPTC}

\begin{figure}
  \includegraphics[width=0.45\textwidth]{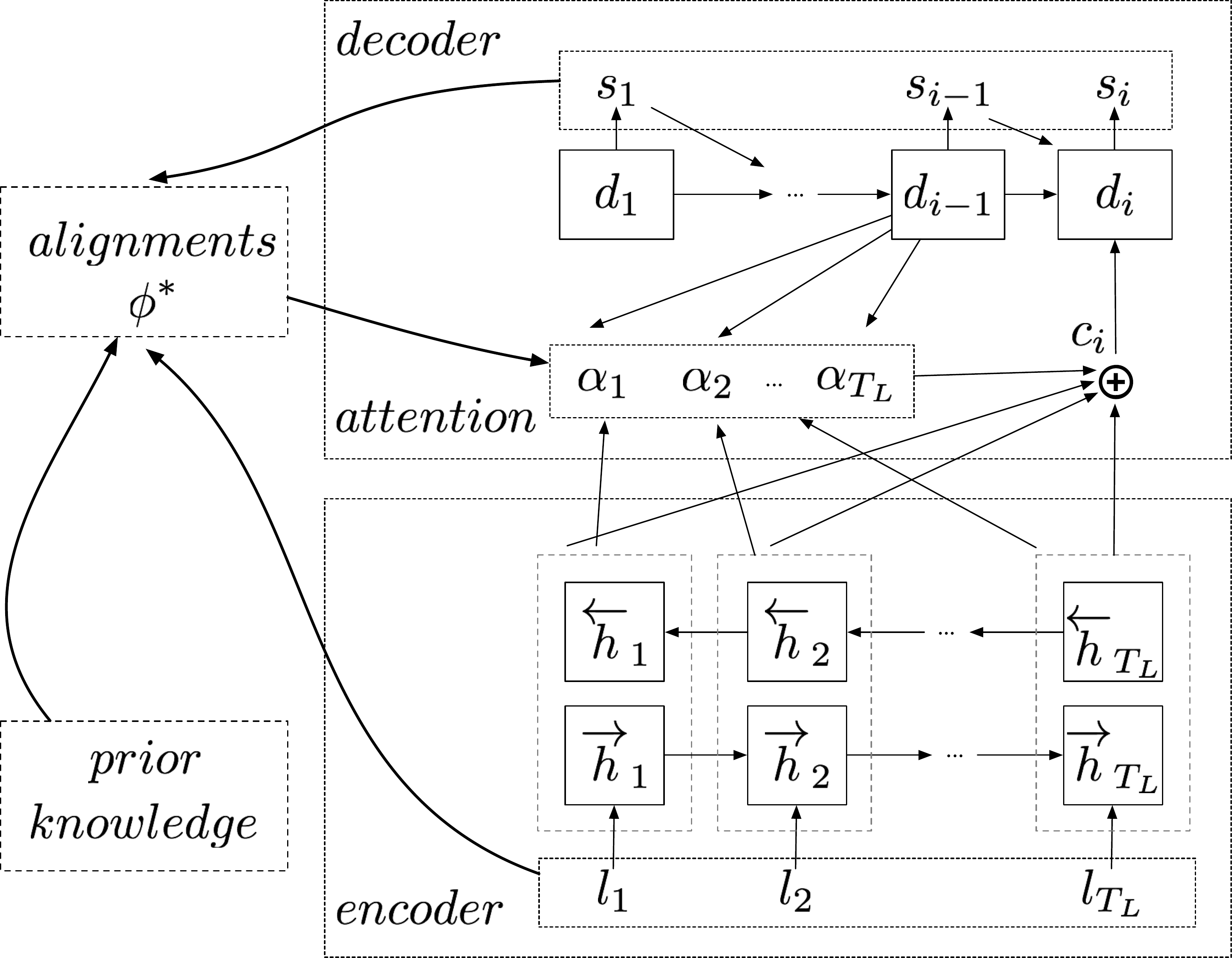}
  \caption{Architecture of the Overall Model}
  \label{fig:arch}
\end{figure}

In this paper, the task of Chinese pinyin typo correction is formulated as a translation task 
, in which the source language $L$ is the user input pinyin sequence $l_1l_2\dots{}l_{T_L}$, where $l_i$ is a letter,
and the target language $S$ is the correct segmented pinyin syllables $s_1s_2\dots{}s_{T_S}$, where $s_i$ is a pinyin syllable.
For example, taking the user's misspelled pinyin input sequence ``nihapma'' as a source sentence, then the corresponding target sentence 
is ``ni hao ma''. 
KNPTC combines implicit representations of pinyin sequence and explicit knowledge of adjacent letters transitions
to further improve the performance of the pinyin typo correction.
Figure~\ref{fig:arch} shows the architecture of KNPTC. 
It consists of a character-level sequence to word-level sequence model as a backbone,
and uses the supervised attention component to jointly learn attention and typo correction.

\subsection{NMT Framework as a Backbone for KNPTC}
The backbone of KNPTC closely follows the basic neural machine translation architecture with attention proposed in ~\cite{DBLP:journals/corr/BahdanauCB14}.
We here present a key description of this model.

The underlying framework of the model is called RNN Encoder–Decoder~\cite{sutskever2014sequence}.
Given a sequence of vectors $\mathbf{x} = (x_1,\dots,x_{T_x})$ representing the source language,
the encoder creates corresponding hidden state vectors $\mathbf{e}$ using a bidirectional RNN~\cite{DBLP:journals/tsp/SchusterP97}:
\begin{equation}
	\mathbf{e} = (h_1,\dots,h_{T_x}).
\end{equation}
The hidden state $h_t$ at time $t$ is obtained by concatenating the forward hidden state $\overrightarrow{h_t}$ and the 
backward one $\overleftarrow{h_t}$, i.e., $h_t = [\overrightarrow{h_t}^{\top}; \overleftarrow{h_t}^{\top}]^{\top}$.
The $\overrightarrow{h_t}$ and $\overleftarrow{h_t}$ are computed as follows: 
\begin{equation}
\begin{split}
	\overrightarrow{h_t} &= \mathrm{GRU}_{\overrightarrow{enc}}(\overrightarrow{h_{t-1}},x_t)\\
	\overleftarrow{h_t} &= \mathrm{GRU}_{\overleftarrow{enc}}(\overleftarrow{h_{t+1}},x_t),
\end{split}
\end{equation}
where $\mathrm{GRU}_{\overrightarrow{enc}}$ and $\mathrm{GRU}_{\overleftarrow{enc}}$ represent gated recurrent unit functions.
The GRUs with subscripts $\overrightarrow{enc}$ and $\overleftarrow{enc}$ indicate the use of two functions with different parameters, respectively. 
Here, $\overrightarrow{enc}$ and $\overleftarrow{enc}$ denote the use of the forward and backward encoder units.

The decoding process also uses the GRU functions, 
which uses a sequence of hidden states $d_1,d_2,\dots,d_{T_y}$ to 
define the probability of the output sequence $y_1,y_2,\dots,y_{T_y}$ as follows:
\begin{equation}
	p(y_i|y_1,\dots,y_{i-1},\mathbf{x}) = g(y_{i-1},d_i,c_i),
\end{equation}
where $g$ is a nonlinear, multi-layered, function that outputs the probability distribution of $y_i$,
and $d_i$ is an RNN hidden state at step $i$, computed as
\begin{equation}
	d_i = \mathrm{GRU}_{dec}(d_{i-1}, y_{i-1}, c_i).
\end{equation}


The context vector $c_i$ used to predict $y_i$ is equal to $\sum_{j=1}^{T_x}{\alpha_{ij}h_j}$.
The weight $\alpha_{ij}$ for $h_j$ is given by:
\begin{equation}
	\alpha_{ij} = \frac{exp(q_{ij})}{\sum_{k=1}^{T_x}{exp(q_{ik})}},\label{eq:weight}
\end{equation}
where $q_{ij} = a(d_{i-1}, h_j)$.
Here $a$ is a feedforward neural network which is jointly trained with the entire network.

Given a training dataset $\mathbb{C}$ containing $|\mathbb{C}|$ pairs of training samples in the form of $<\mathbf{x},\mathbf{y}>$, 
the training goal of the model is to minimize the cross entropy loss as follows:

\begin{equation}
	Loss' = -\sum_{<\mathbf{x},\mathbf{y}>\in\mathbb{C}}\sum_{l=1}^{T_y}\log{p(y_l|y_{<l},\mathbf{x})}\label{eq:lossFunc}.
\end{equation}

\subsection{Supervised Attention Model with Keyboard Neighborhoods Information}

The attention $\alpha_{i,1}, \alpha_{i,2}, \dots, \alpha_{i, T_{x}}$ in each step plays an important role in predicting the next word.
However, as shown in equation~\eqref{eq:weight}, 
the original attention mechanism, which selectively focuses on the source words, does not take into account any knowledge of the keyboard neighborhoods, which is crucial for Chinese pinyin typo correction. 
This knowledge is especially useful when tackling typo correction for acronym pinyin input.
Thus, in this section, we introduce discrete and probabilistic keyboard neighborhoods information 
of users pinyin input 
to capture the explicit knowledge about users input behaviors 
for enhancing the attention mechanism and improving pinyin typo correction quality.
To integrate the keyboard neighborhoods information into attentional NMT,
we propose a novel alignment model using the transition probabilities between adjacent letters.
We then use this alignment model to supervise the learning of the attention mechanism.

Assuming that we have such a prior knowledge $p_t$, given a letter $l_i$, we assign a probability $p_t(l_i\rightarrow{}l_s)$ for any other letter $l_s$.
$p_t(l_i\rightarrow{}l_s)$ represents the probability that the user wants to enter $l_i$ but type in $l_s$ instead since $l_s$ is adjacent to $l_i$ on the keyboard, i.e., $p_t(l_i\rightarrow{}l_s)$ is the probability that letter $l_i$ transfers to the letter $l_s$. 
Given a vocabulary $V_l$ containing letters on the keyboard~(for instance, 26 Latin letters),
for a letter $l_s$, the probability $p_t(l_i\rightarrow{}l_s)$ will be zero for most letters in $V_l$.
We first describe how to integrate these probabilities into attentional NMT, 
and then explain how to obtain the $p_t$ from the dataset.

\subsubsection{Converting Transition Probabilities into Alignment Model}
\makeatletter
\newcases{crcases}{\quad}{%
  \hfil$\m@th\displaystyle{##}$\hfil}{\hfil$\m@th\displaystyle{##}$}{\lbrace}{.}
\makeatother
For each pinyin sequence pair $(L,S)$, we define an alignment matrix $\phi$ with $T_L + 1$ rows($T_L$ letters and one <EOS>) and $T_S + 1$ columns($T_S$ pinyin words and one <EOS>).
For $1 \leq i \leq T_L$ and $1 \leq j \leq T_S$, $\phi_{ij}$ is computed as:
\begin{equation}
	\phi_{ij} = \max_{\substack{1 \leq k \leq |S_j|}}{p_t(s_{jk}\rightarrow{}l_i)A_{i,j}},
	\label{eq:phi-ij}
\end{equation}
and for $ i = T_L+1 $ or $ j = T_S+1 $, $\phi_{ij}$ is computed as:
\begin{equation}
	\phi_{ij} = 
	\begin{crcases}
		1 & i = T_L + 1, j = T_S + 1 \\
		0 & \text{otherwise}
	\end{crcases}\label{eq:fai-ij-other},
\end{equation}
where $A_{ij} = 1$ in equation~\ref{eq:phi-ij} denotes that letter $l_i$ is translated as a part of word $S_j$, otherwise $A_{ij} = 0$.
In fact, $A$ represents a segmentation of the input sequence.
Given input sequence $L$ and pinyin words $S$, algorithm~\ref{algo: genSegmentation} shows how to obtain the segmentation of $L$. 
In line 29 of algorithm~\ref{algo: genSegmentation}, we notice that there is a procedure \textsc{CalcScore}.
Supposing that variable \textit{Seg} contains a segmentation of $L$ which is represented as $\{seg_1,seg_2,\cdots,seg_{T_S}\}$,
we then use the procedure \textsc{CalcScore} and the edit distance~\cite{DBLP:journals/jacm/WagnerF74} to compute the score $\zeta$ of this segmentation by:

\begin{equation}
	\zeta = -\sum_{n = 1}^{T_S}{\textit{EditDistance}(seg_n, s_n)}.
\end{equation}


\makeatletter
\def\BState{\State\hskip-\ALG@thistlm}
\makeatother

\begin{algorithm}[h]
\caption{Generating segmentation of the input sequence $L$}\label{algo: genSegmentation}
\begin{algorithmic}[1]

\Variables
 \State \textit{SEGMENTATIONS} \Comment Candidate segmentations
 \State \textit{SCORES} \Comment Scores for candidate segmentations
 \State \textit{MAX\_LEN} \Comment Max length of Chinese pinyin words
\EndVariables

\Procedure{GenSegmentation}{$L$,$S$}

\State \textit{SegNum} $\gets$ $T_S - 1$
\State \text{Initialize}~\textit{Seg}~\text{as an empty string}
\State \textit{SCORES} $\gets$ [~]
\State \textit{SEGMENTATIONS} $\gets$ [~]
\State \textit{BeginPos} $\gets$ 0
\State \textit{MAX\_LEN} $\gets$ 6 \Comment{6 is the max length of Chinese pinyin words}

\State \Call{Segment}{\textit{BeginPos, SegNum, Seg, L, S}}

\State \textit{MaxScore} $\gets$ \Call{Max}{\textit{SCORES}}
\State \textit{index} $\gets$ $\argmax_{index}{\textit{SCORES}}$
\State \Return \textit{SEGMENTATIONS}[\textit{index}]
\EndProcedure

\Procedure{Segment}{\textit{BeginPos},\textit{SegNum},\textit{Seg},$L$,$S$}

\If {\textit{SegNum} $> 0$}
	\If {\textit{BeginPos} $+$ \textit{MAX\_LEN} $<T_L$}
		\State \textit{EndPos} $\gets$ \textit{BeginPos} $+$ \textit{MAX\_LEN}
	\Else
		\State \textit{EndPos} $\gets{} T_L$ 
	\EndIf
	\For{$i\gets$ \textit{BeginPos}$+1$ to \textit{EndPos}}
		\State \textit{Seg} $\gets$ \textit{Seg} $+$ \textit{L}[\textit{BeginPos, i~}] $+$ ``' ''
		\State \textit{SegNum} $\gets$ \textit{SegNum} $-1$
		\State \Call{Segment}{\textit{i, SegNum, Seg, L, S}}
	\EndFor
	\Else
	\State \textit{Seg} $\gets$ \textit{Seg} $+$ \textit{L}[\textit{BeginPos~}:]
	\State $\zeta$ $\gets$ \Call{CalcScore}{\textit{Seg, S}}
	\State Append $\zeta$ to \textit{SCORES}
	\State Append \textit{Seg} to \textit{SEGMENTATIONS}
\EndIf

\EndProcedure

\end{algorithmic}
\end{algorithm}

\begin{figure}[h!]
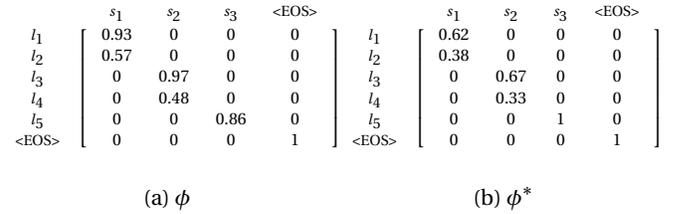

    \centering
    \begin{subfigure}[b]{0.23\textwidth}
    	\centering
		\renewcommand{\kbldelim}{[}
		\renewcommand{\kbrdelim}{]}
		\scriptsize
		\[
    		\kbordermatrix
    		{
    			& s_1 & s_2 & s_3 & \text{\tiny{<EOS>}} \\
    			l_1 & 0.93 & 0 & 0 & 0  \\
    			l_2 & 0.57 & 0 & 0 & 0  \\
    			l_3 & 0 & 0.97 & 0 & 0  \\
    			l_4 & 0 & 0.48 & 0 & 0  \\
    			l_5 & 0 & 0 & 0.86 & 0  \\
    	\text{\tiny{<EOS>}}& 0 & 0 & 0 & 1  \\
    		}
		\]
		\normalsize
		\caption{$\phi$}
		\label{fig:fai}
    \end{subfigure}\hfill
    \begin{subfigure}[b]{0.23\textwidth}
    	\centering
		\renewcommand{\kbldelim}{[}
		\renewcommand{\kbrdelim}{]}
		\scriptsize
		\[
    		\kbordermatrix
    		{
    			& s_1 & s_2 & s_3 & \text{\tiny{<EOS>}} \\
    			l_1 & 0.62 & 0 & 0 & 0  \\
    			l_2 & 0.38 & 0 & 0 & 0  \\
    			l_3 & 0 & 0.67 & 0 & 0  \\
    			l_4 & 0 & 0.33 & 0 & 0  \\
    			l_5 & 0 & 0 & 1 & 0  \\
    	\text{\tiny{<EOS>}}& 0 & 0 & 0 & 1  \\
    		}
		\]
		\normalsize
		\caption{$\phi^*$}
		\label{fig:fai star}
    \end{subfigure}


    \caption[]{
		The original alignment matrix $\phi$ and the alignment matrix $\phi^*$ after transformation.
	}
    \label{fig:alignment mat}
    
\end{figure}

Given an original alignment matrix $\phi$ in Figure~\ref{fig:fai}, we use scaling transformation to get the probability distribution $\phi^*$.
Figure~\ref{fig:fai star} shows the result matrix $\phi^*$.
We then use $\phi^*$ to supervise the learning of the attention mechanism.



%

\subsubsection{Using Alignment Model to Supervise Attentional NMT}

Based on the ``ground-truth'' alignment matrix $\phi^*$ mentioned above and the attentions $\phi'$ generated by the NMT model, 
we define the following distance function between $\phi^*$ and $\phi'$:

\begin{equation}
	\ell(\phi^*, \phi') = \|\phi^* - \phi'\|_2^2 \label{dist func}.
\end{equation}

Combined with the previous function in equation~\ref{eq:lossFunc}, we finally obtain the following loss function:

\begin{equation}
	Loss = -\sum_{<\mathbf{x},\mathbf{y}>\in\mathbb{C}}\left\{\sum_{l=1}^{T_y}\log{p(y_l|y_{<l},\mathbf{x})} - \ell(\phi^*, \phi') \right\}. \label{eq:finalLossFunc}
\end{equation}

As shown above, the loss function consists of two parts. 
The first part represents the loss term about typo correction and 
the second part represents the loss term about alignment model which incorporates the transition probabilities between adjacent letters.
We then minimize them jointly in our training process.

\subsubsection{Obtaining Prior Knowledge $p_t$ from Data}
In the previous sections, we propose a method of using prior knowledge $p_t$ to supervise the learning of attention mechanism.
Now, we define the ways to obtain the prior knowledge $p_t$ from the real-life input behaviors of anonymous users.

$p_t(l_i\rightarrow{}l_j)$ represents the probability that the user wants to enter $l_i$ but type in $l_j$ instead, since $l_j$ is adjacent to $l_i$ on the keyboard.
These probabilities can be learned directly from users' editing actions.
Specifically, users press backspace on the keyboard to modify the typos, 
then they delete the mistyped letters and re-type in the correct ones.
Motivated by this observation, $p_t(l_i\rightarrow{}l_j)$ is calculated by:

\begin{equation}
	p_t(l_i\rightarrow{}l_j) = \frac{N(l_i\rightarrow{}l_j)}{N(l_i)},
\end{equation}
where $N(l_i\rightarrow{}l_j)$ denotes the number of times that the letter $l_i$ is wrongly entered as the letter $l_j$ in the actual editing actions, and $N(l_i)$ indicates the total number of times the letter $l_i$ being inputted.

%


\section{Experiments}


\subsection{Data Preparation}

Our real-life datasets come from user logs of the Chinese input method called Sogou-Pinyin\footnote{\url{https://pinyin.sogou.com}}.
The user logs mainly include the following parts:

\begin{itemize}
	\item The user's input string, which is an unsegmented sequence of Latin letters (henceforth referred to as ``original input'');
	\item The desired Chinese sentence selected by the user (henceforth referred to as ``target Chinese sentence'') from the candidates,
	which are generated by the IME and consist of one or more Chinese words;
	\item The backspace and re-enter operations made by the user will be recorded if the candidates do not meet the user's needs due to typos or other reasons.
\end{itemize}

We extracted original input and target Chinese sentence pairs from these data for experiments.
Under the normal circumstances, users should input the complete pinyin sequence corresponding to the target Chinese sentence.
However, original input may contain abnormal input due to typos.
We classified abnormal input into two types: acronym pinyin input and misspelled pinyin input~(acronym pinyin input also contains typos).
According to the acronym pinyin's position in the original input, we simply divided acronym pinyin input into two categories:
\begin{itemize}
	\item \textbf{Global acronym pinyin input}.
	Each of the Chinese characters uses acronym pinyin for representing, for example: ``你好吗$\rightarrow${}~ni hao ma$\rightarrow$~n h m''.
	\item \textbf{Local acronym pinyin input}.
	Only a part of Chinese characters use acronym pinyin for representing, for example: ``你好吗$\rightarrow${}~ni hao ma$\rightarrow$~ni hao m'' or ``你好吗$\rightarrow${}~ni hao ma$\rightarrow$~ni h m''.
\end{itemize}

\begin{table}[h]
\centering
\begin{tabular}{|c|c|c|}
\hline
\thead{\textbf{User input}} & \thead{\textbf{Target Chinese sentence} \\ \textbf{(in pinyin form )}}  & \textbf{Input type}         \\ \hline
dadianhuageini             & da~ dian~ hua~ gei~ ni                                            & CP    \\ \hline
xianzqub               & xian~ zai~ qu~ ba                                            & LAP   \\ \hline
yijianzq               & yi~ jian~ zhong~ qing                                         & LAP   \\ \hline
bgnll              & bu~ gen~ ni~ liao~ le                                  & GAP   \\ \hline
niyiubudhi              & ni~ you~ bu~ shi                                             & MP    \\ 
\hline
\end{tabular}

\caption{
	Some sample data extracted from user logs and its corresponding input type.
	For convenience, we use \textbf{CP} to indicate the correct pinyin input,
	\textbf{LAP} to be the local acronym pinyin input, 
	\textbf{GAP} for the global acronym pinyin input and \textbf{MP} for the misspelled pinyin input.
}
\label{tbl: example data}
\end{table}

\begin{table}[h]
\centering
\begin{tabular}{|c|c|c|}
\hline
\textbf{Input type}           & \textbf{\#Records} & \textbf{Proportion(\%)} \\ \hline
CP                 & 488,583            & 41.58                           \\ \hline
LAP           & 408,228            & 34.74                           \\ \hline
GAP          & 211,261            & 17.98                           \\ \hline
MP          & 66,717             & 5.70                            \\

\hline
\end{tabular}
\caption{Input types distribution. We can observe that there is a large proportion of acronym pinyin input in the real-life data.}
\label{tbl: records number}
\end{table}

Table~\ref{tbl: example data} shows examples that we extracted from the user logs.
The distribution of the input types is given in Table~\ref{tbl: records number}.
It should be noted that the original input is a continuous sequence without segmentation . 
In addition, unlike English and other languages, the original input only expresses the pronunciation information of the target Chinese sentence, 
so we translated the target Chinese sentence into its pinyin form to obtain the target pinyin sentence.
We treat the original input as the source sentence and the target pinyin sentence as the target sentence in our approach.
Finally, 1,179,789 anonymous users' typing records during 2 days were collected in total.
From the extracted dataset with about 1M parallel sentences, 
we randomly selected close to 50K parallel sentences for using as a development set, 100K for testing and 850K for training.

\begin{table*}[t]
\centering
\begin{tabular}{|c|c|c|c|c|}
\hline
\textbf{Original input} & \textbf{Target sentence} & \textbf{JGM}     & \textbf{GoogleIT}   & \textbf{KNPTC}     \\ \hline
yijiam                  & yi ~jian   & yi~~jian          & yi~ jia~ men          & yi~ \textbf{jian}                         \\ \hline
yigeciaos               & yi~ ge~ xiao~ shi      & yi~~ge~~xiao~~a     & yi~ ge~ ci~ ao~ shen   & yi~ ge~ \textbf{xiao}~ \textbf{shi}       \\ \hline
zhongqiykuail           & zhong~ qiu~ kuai~ le      & zhong~~qiu~~kuai~~a & zhong~ qi~ yi~ kuai~ le & zhong~ \textbf{qiu}~ kuai~ \textbf{le} \\ \hline
smshuh                  & shen~ me~ shi~ hou        & si~~shuo          & shuo ~ming ~shu~ he    & \textbf{shen~ me} ~\textbf{shi}~ \textbf{hou}    \\ \hline
gabgshuix               & gang~ shui~ xing       & gang ~shui~ a      & ga~ bu~ guo~ shui~ xing  & \textbf{gang}~ shui~ \textbf{xing}       \\ \hline
\end{tabular}
\caption{
	Example typo corrections made by JGM, GoogleIT and KNPTC.
	Typo correction made by KNPTC is in \textbf{bold}.	
}
\label{tbl:typo-example}
\end{table*}

\subsection{Evaluation Metrics and Baselines}
We evaluate KNPTC with conventional sequence labeling evaluation metrics: word accuracy and sentence accuracy, 
which have significant impacts on user experience of IMEs~\cite{jia2014joint}.

We compare KNPTC with two approaches: GoogleIT and JGM~(state-of-the-art method).
GoogleIT contains a practical pinyin input method.
We use its public APIs for testing.
JGM is proposed in~\cite{jia2014joint}.
The basic idea of JGM is to adopt the graph model for pinyin typo correction and segmentation jointly.
We obtain its source code from the author\footnote{\url{http://bcmi.sjtu.edu.cn/~zhaohai}} and use its default settings for hyper parameters.
The metrics for evaluating each model are the accuracy of prediction.




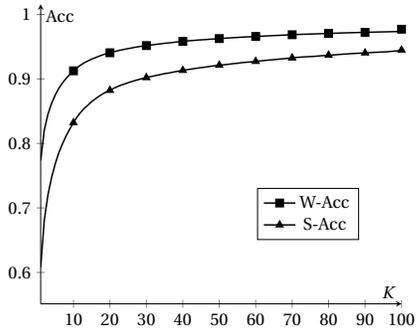
\begin{figure}[h]
    	\centering
		\begin{tikzpicture}[scale=0.70]
    			\begin{axis}[
    			axis lines=middle,
    			ymin=0.59,
    			xlabel=$K$,
    			ylabel=Acc,
    			enlargelimits = false,
    			enlarge y limits = true,
    			xtick={0,10,20,...,100},
				samples=10,
    			legend style={at={(0.6,0.3)},anchor=west}
			]
			\addplot[black,thick,mark=square*,mark repeat=10,mark phase=10] table [y=W-Acc,x=X]{KNPTC-Acc-100.dat};
			\addlegendentry{W-Acc}
			\addplot[black,thick,mark=triangle*,mark repeat=10,mark phase=10] table [y=S-Acc,x=X]{KNPTC-Acc-100.dat};
			\addlegendentry{S-Acc}
			\end{axis}
		\end{tikzpicture}
    \caption[]{W-Acc and S-Acc with different $K$s for KNPTC.}
    \label{fig:acc-compare}
\end{figure}

\begin{table}[h]
\centering
\begin{tabular}{|c|c|c|c|c|}
\hline
\multicolumn{1}{|c|}{\textbf{Input type}} & \textbf{Acc} & \textbf{JGM} & \multicolumn{1}{c|}{\textbf{GoogleIT}}  & \multicolumn{1}{c|}{\textbf{KNPTC}} \\ \hline
\multirow{2}{*}{CP}    
                       & W-Acc   & 99.86          &99.89                        & 99.83                      \\ \cline{2-5} 
                       & S-Acc   & 99.51          &99.78                        & 99.64                      \\ \hline
\multirow{2}{*}{LAP}  
                       & W-Acc   & 55.06          &90.19                        & \textbf{93.11}                      \\ \cline{2-5} 
                       & S-Acc   & 0              &78.20                        & \textbf{83.40}                      \\ \hline
\multirow{2}{*}{GAP}   
                       & W-Acc   & 17.01          &61.78                        & \textbf{62.94}                      \\ \cline{2-5} 
                       & S-Acc   & 0              &35.77                        & 31.18                      \\ \hline
\multirow{2}{*}{MP}    
                       & W-Acc   & 68.11          &50.90                        & \textbf{95.19}                      \\ \cline{2-5} 
                       & S-Acc   & 47.84          &11.58                        & \textbf{88.87}                      \\ \hline
\multirow{2}{*}{MIX}    
                       & W-Acc   & 68.69          &84.09                        & \textbf{91.25}                      \\ \cline{2-5} 
                       & S-Acc   & 44.85          &70.58                        & \textbf{83.22}                      \\ \hline
\end{tabular}

\caption{
	Comparison with baseline systems on identical dataset for different input types.
	MIX stands for CP+LAP+GAP+MP.
}
\label{tbl:result}
\end{table}

\subsection{Training Details and Results}
The embedding size of character-level encoder and word-level decoder is set to 256, and the hidden unit size is 128.
All the parameters are initialized from a uniform distribution. 
Our model converges after about 350K iterations.
When running on a single GPU device Tesla K40, it takes two days to train our model.
We choose $K$ best candidates for pinyin typo correction.
Figure~\ref{fig:acc-compare} shows the results of KNPTC on development dataset for different $K$s.
We choose the value of $K$ when the improvements of performance on W-Acc of $K$ best candidates is less than the threshold $\tau$.
Considering the practical need for the number of candidates in IMEs,
we set $\tau$ to 0.005 and get $K=10$ for pinyin typo correction.

The examples of typo corrections in Table~\ref{tbl:typo-example} demonstrate the abilities of KNPTC,
which is able to handle a variety of pinyin typos, 
e.g., due to keyboard neighborhood~(\textit{qiy} $\rightarrow$ \textit{qiu}, \textit{gabg} $\rightarrow$ \textit{gang}) or
acronym pinyin~(\textit{smshuh} $\rightarrow$ \textit{shen me shi hou}).
Considering that the users' real-life input contains different types, 
to further inspect the robustness of KNPTC,
we evaluate the performance of each method separately on all input types, i.e., CP, LAP, GAP, MP and mixed input type.
Table~\ref{tbl:result} shows the performance of JGM, GoogleIT and KNPTC on the test dataset.
Some observations can be derived:
\textbf{1)}~KNPTC performs significantly better than JGM in almost all the cases,
especially for LAP and GAP(with 38.05\% and 45.93\% improvements for W-Acc).
From Table~\ref{tbl:typo-example} we notice that JGM fails to generate the correct sentence for acronym pinyin input due to its shortness and sparsity, 
while KNPTC performs well on both LAP and GAP.
The reason is that KNPTC is more capable of learning the sensible semantic information and typing patterns of user input.
Besides, KNPTC is able to capture the alignments information with transition probabilities between adjacent letters  
in the decoding stage, and thus it can select more relevant input to predict the next target word.
It is worth noting that acronym pinyin input occupies a large part in real-life user input.
Therefore, KNPTC is more practical than JGM.
\textbf{2)}~KNPTC outperforms GoogleIT, 
especially for MP(with 44.29\% and 77.29\% improvements for W-Acc and S-Acc).
From Table~\ref{tbl:typo-example} we can observe that GoogleIT fails to correctly segment the 
original pinyin input due to the existence of typos and therefore it can not produce correct results.
KNPTC overcomes this problem by finding the optimal segmentation and typo correction jointly on the user's original input pinyin sequence.
We can also observe that 
KNPTC's S-Acc on GAP is slightly lower than that of GoogleIT.
This can be attributed to the fact that 
the language model of Chinese sentences is not been fully exploited in KNPTC, 
which is useful for candidates generation.
To make typo correction better, we plan to integrate it with KNPTC in future work.
\textbf{3)}~KNPTC's S-Acc on LAP, MP and MIX is significantly better than that of GoogleIT and JGM.
This means that KNPTC can generate better candidate sentences, 
which is critical to the user experience of IMEs.
From the results in Table~\ref{tbl:result}, we can further conclude that 
the alignments information with transition probabilities between adjacent letters can help improve the overall performance.

\section{Conclusion}
In this paper, we propose an effective approach called KNPTC to integrate the transition probabilities between adjacent letters into attentional NMT to capture more sensible typo correction patterns.
KNPTC finds the optimal segmentation and typo correction jointly on the user's original input pinyin sequence.
In addition, KNPTC can also cope with the typo correction of acronym pinyin input.
Experiments show that KNPTC can evidently improve the pinyin typo correction performance.
To our best knowledge, 
our work is among the earliest studies in leveraging neural machine translation for Chinese pinyin typo correction.


\bibliographystyle{named}

\end{CJK}
\end{document}